\documentclass[letterpaper, 10 pt, conference]{ieeeconf}  

\IEEEoverridecommandlockouts                              

\usepackage{amsmath} 
\usepackage{amssymb}  
\usepackage{graphicx}
\usepackage{color}
\usepackage{subcaption}
\usepackage{multirow}
\usepackage[colorinlistoftodos]{todonotes}

\usepackage{bbm} 

\newcommand{\filmbox}[1]{
\colorbox{black}{%
\begin{minipage}{3.2cm}
\rule{0mm}{0mm}\null\\%
\null\hfill\includegraphics[width=3.2cm,trim={2.1cm 3.6cm 2cm 3.6cm},clip]{#1}\hfill
\end{minipage}}}

\title{\LARGE \bf
Interaction-Aware Multi-Agent Reinforcement Learning for Mobile Agents with Individual Goals
}

\author{Anahita Mohseni-Kabir$^{1}$, David Isele$^{2}$, and Kikuo Fujimura$^{2}$
\thanks{$^{1}$Anahita Mohseni-Kabir is with the  Robotics  Institute,  Carnegie  Mellon
University, Pittsburgh, PA 15213, USA {\tt\small anahitam@andrew.cmu.edu}. This work has been conducted while she was at Honda Research Institute USA.}%
\thanks{$^{2}$David Isele and Kikuo Fujimura are with Honda Research Institute USA
        {\tt\small \{disele, kfujimura\}@honda-ri.com}}%
}

\begin{document}

\maketitle
\thispagestyle{empty}
\pagestyle{empty}
\begin{abstract}
In a multi-agent setting, the optimal policy of a single agent is largely dependent on the behavior of other agents. We investigate the problem of multi-agent reinforcement learning, focusing on decentralized learning in non-stationary domains for mobile robot navigation. We identify a cause for the difficulty in training non-stationary policies: mutual adaptation to sub-optimal behaviors, and we use this to motivate a curriculum-based strategy for learning interactive policies. The curriculum has two stages. First, the agent leverages policy gradient algorithms to learn a policy that is capable of achieving multiple goals. Second, the agent learns a modifier policy to learn how to interact with other agents in a multi-agent setting. 
We evaluated our approach on both an autonomous driving lane-change domain and a robot navigation domain. 
\end{abstract}

\section{Introduction}


Single agent reinforcement learning (RL) algorithms have made significant progress in game playing~\cite{mnih2015human} and robotics~\cite{kober2013reinforcement}, however, single agent learning algorithms in multi-agent settings are prone to learn stereotyped behaviors that over-fit to the training environment
\cite{raghu2017can,lanctot2017unified}. 
There are several reasons why multi-agent environments are more difficult: 1) interacting with an unknown agent requires having either multiple responses to a given situation or a more nuanced ability to perceive differences. The former breaks the Markov assumption, the latter rules out simpler solutions which are likely to be found first. 2) Intentions and goals of other agents are not known and must be inferred. This also can break the Markov assumption. 3) Agents are co-evolving, and their policies are non-stationary during training. 
In this work we investigate a property associated with non-stationary policies: partially successful policies for one agent can get repeated multiple times or 'burned in', causing the other agents to adapt to that specific behavior. This sets off a chain of mutual adaptation that encourages agents to only visit suboptimal regions of the state space.

\begin{figure} [tb]
\centering
  \includegraphics[width=0.8\linewidth]{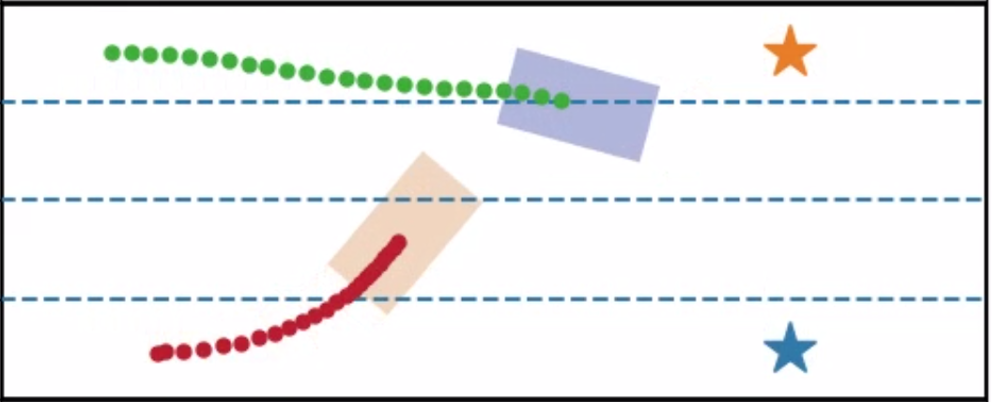}
  \caption{Double lane-change problem. The bottom car and the top car are crossing one another to go to the star on the top right and bottom right of the environment respectively.}
  \label{fig:simulator}
\end{figure}

When training independent agents~\cite{tan1993multi}, the competing learning processes of multiple agents is sufficiently difficult that either the agents fail to learn, or the agents learn one-after-the-other, resulting in stereotyped policies that are sensitive to the behavior of other agents. Recent approaches to multi-agent reinforcement learning have relaxed the independence of an agent by exploiting centralized training and assuming the other agents' actions are known \cite{foerster2017counterfactual,lowe2017multi}. While these techniques are more successful, we show that they still exhibit one-after-the-other learning, producing highly-dependent policies. 

We propose the use of an independent curriculum-based training procedure for learning policies that avoids mutual adaptation without using either centralized training or knowledge of the other agents actions.  
We start from the observation that when agents learn at different rates (as often happens under random initialization), one agent learns a policy \emph{around} the other agents' policies. Since the policies being accommodated are suboptimal during the early stages of learning, the agents drive each other into poor local optima. One solution to pacing the learning of multiple agents is self-play~\cite{silver2016mastering}, but self-play requires symmetric agents. As an alternative, we consider learning via a curriculum~\cite{bengio2009curriculum}. The use of curriculum learning lets us pace the learning of each agent, while allowing us to handle the case of agents with different goals. 

We structure our curriculum by first learning an optimal single agent and then explicitly learning the modifications required to interact with other agents. 
This approach leverages the intuition that when other agents are not present in the environment, the agent should behave like a single agent trying to reach a goal. 
We address the interaction-aware decision making problem in the second stage of the curriculum. 
In the second stage, we introduce an architecture that uses the learned single-agent policy, and adjusts it with a learned interactive multi-agent policy. 

We consider two robotic applications where the agents’ policies must be mutually consistent in order to achieve the intended goals: a lane-change scenario and a mobile navigation scenario. 
The agents do not have access to the goals or intentions of other agents and are learning different policies simultaneously. We show that not only is our curriculum-based approach better able to learn the desired behaviors, but that the learned policies also generalize against agents that were not included in the training process.

\section{Related Work}
\noindent Multiple works have focused on reinforcement learning methods for multi-agent domains in fully cooperative, fully competitive, and mixed environments~\cite{busoniu2008comprehensive}. Foerster et al. ~\cite{foerster2017counterfactual} propose a multi-agent actor-critic method to address the challenges of multi-agent credit assignment. Different from our work where each agent pursues its individual goal, their approach is appropriate for problems with a single shared task. Lowe et al. ~\cite{lowe2017multi} propose an actor-critic approach, called MADDPG, that augments the agent's critic with action policies of other agents 
and is able to successfully learn the coordination policy between multiple agents. Unlike our approach these methods use centralized training.

He et al.~\cite{he2016opponent} presents new models based on deep Q-network for decision-making in adversarial games which  jointly learns a policy and different strategies of opponents. Their proposed approach is appropriate for two agents, whereas our approach is tested on more than two agents. Furthermore, our approach is demonstrated in the non-stationary case where the agents are co-evolving together while the opponents in their approach have a set of fixed strategies. Some work leverages self-play to provide the agents with a curriculum for learning complex competitive tasks~\cite{bansal2017emergent}. They use dense rewards in the beginning phase of the training to allow the agents to learn basic motor skills. In our environments, it is difficult to create motor skills or define a reward function for them. In our problems, self-play using single agent RL algorithms failed to learn successful policies since the environment is non-stationary, and also without cooperation the agents were not able to even occasionally succeed in their tasks thus were not able to learn successful policies.

In social dilemma research, most works focus on one-shot or repeated tasks~\cite{mathieu2015new,hao2015introducing} and ignore that in real world scenarios the behaviors are temporally extended. Among the most relevant RL approaches in this area, is work on the sequential prisoner's dilemma (SPD)~\cite{leibo2017multi} which leverages deep Q-networks to study effects of environmental parameters on the agents’ degree of cooperation. In contrast to this work that focuses on the impact of the games' parameters on the agents' behavior, we provide a multi-agent learning approach for cooperative problems with individual goals. Another relevant work~\cite{wang2018towards} proposes a deep RL approach for mutual cooperation in SPD games. Their approach adaptively selects its policy with the proper cooperation degree based on the detected cooperation degree of an opponent. Different from their approach, our approach is not specific to two player games. In addition, since we focus on mobile navigation problems our approach learns how to react to the continuous policies of other agents, not just their cooperation degree. 

Significant amount of work has focused on motion planning for autonomous vehicles~\cite{paden2016survey} where the problem 
of intention prediction or trajectory estimation of other agents has been studied. Among these, relevant work has focused on intention-aware POMDP planning for autonomous vehicles~\cite{bai2015intention}. These methods leverage machine learning methods to learn models of other agents as also surveyed in~\cite{albrecht2018autonomous}. In contrast to these approaches, we focus on RL algorithms for interaction-aware agents where the agents are co-evolving together and their motion models are dependent on others policies. Another work ~\cite{bhattacharya2010multi} presents an approach for computing optimal trajectories for multiple robots in a distributed fashion.

\section{Approach}
\noindent We focus on multi-goal multi-agent settings, 
where each agent cooperates with other agents in order to accomplish their individual goals. We leverage the intuition that in many settings, like autonomous driving, the interaction between multiple agents is limited to certain parts of the state space where conflict of interest is present, otherwise the agent behaves according to its single-agent policy. \emph{I.e.}, the agent starts with its own single-agent policy and adapts it to account for the multiple agents that appear in the environment.
We propose a two stages approach to learn multiple interactive policies for multiple agents. In the first stage of learning, the agents learn a single agent policy to accomplish their individual goals. The learned single agent policies are then passed to the multi-agent model that enables each agent to learn an interactive policy to account for the other agents. 

\subsection{Single Agent Module}


\noindent We model an agent with individual goal as a Markov Decision Process (MDP)~\cite{white1989markov} with goals \cite{schaul2015universal}. The MDP is defined as 
a tuple $<\mathcal{S},\mathcal{O},\mathcal{A},P,R, G, \mathcal{G},\gamma>$ in which $\mathcal{S}$ represents the state of the world, $\mathcal{O}$ represents the agent's observation, $\mathcal{A}$ is a set of actions, $P:\mathcal{S} \times \mathcal{A} \rightarrow \mathcal{S}$  determines the distribution over next states, G is the agent's goal, $\mathcal{G}$ is the goal distribution, $R: \mathcal{S} \times \mathcal{G} \times \mathcal{A} \rightarrow \mathbb{R}$ is an immediate reward function, and $\gamma \in [0,1]$ is the discount factor. The solution to an MDP is a policy $\pi_{\theta}: \mathcal{O} \times \mathcal{G} \times \mathcal{A} \rightarrow [0,1]$ where $\theta$ is the parameters of the policy. For continuous actions, $\pi_{\theta}$ is assumed to be Gaussian, and in our work the mean is represented by a neural network with parameters $\theta$.
The robot seeks to find a policy $\pi_{\theta}$ that maximizes the expected future discounted reward $R=\sum_{t=0}^{T} \gamma^{t}r_t$. In the following paragraphs, we add subscript ``self" or ``s" to our notation to show the agent's own properties and ``single" or "sng" to highlight the single agent scenario. 

We use a decentralized actor-critic policy gradient algorithm to learn the single agent policies. The single agent model gets observation $o_{s}$ and goal $g_{s}$ as inputs, and outputs an action $a_{s}$ that the agent should take. Each agent learns a policy $\pi_{sng}(a_{s}|o_{s},g_{s})$, according to its individual goal-specific reward function $R_{sng}(s_{s},a_{s},g_{s})$ in the absence of other agents. The decentralized actor-critic policy gradient algorithm maximizes $J_{sng}(\theta)=\mathop{\mathbb{E}}_{o_{s} \sim p_{s}^\pi,a_{s} \sim {\pi_{sng}},g_{s} \sim \mathcal{G}}{[R_{sng}]}$ by ascending the following gradient:
\begin{multline*}
\nabla_{\theta} {J(\theta)} = \mathop{\mathbb{E}}_{\tau \sim p_{\theta}(\tau), g_{s} \sim \mathcal{G}}  [\nabla_{\theta} \log {\pi_{sng}(a_{s}|o_{s},g_{s})} \\
 A^{\pi}(o_{s},a_{s},g_{s})]
\end{multline*}
Where $p_{s}^\pi$ is the state distribution, and $A^{\pi}$ is the advantage function \cite{schulman2015trust}. 
We use $\tau \sim p_{\theta}(\tau)$ to refer to $o_{s} \sim p_{s}^\pi,a_{s} \sim {{\pi}_{sng}}$. For simplicity, $\theta$ is removed.
The gray colored boxes in Fig.~\ref{fig:multi_model} show the actor-critic model for the single agent.

\subsection{Multiple Agents Module}

\begin{figure} [tb]
\centering
  \includegraphics[width=\linewidth,trim={0.4cm 0.3cm 0.25cm 0.4cm},clip]{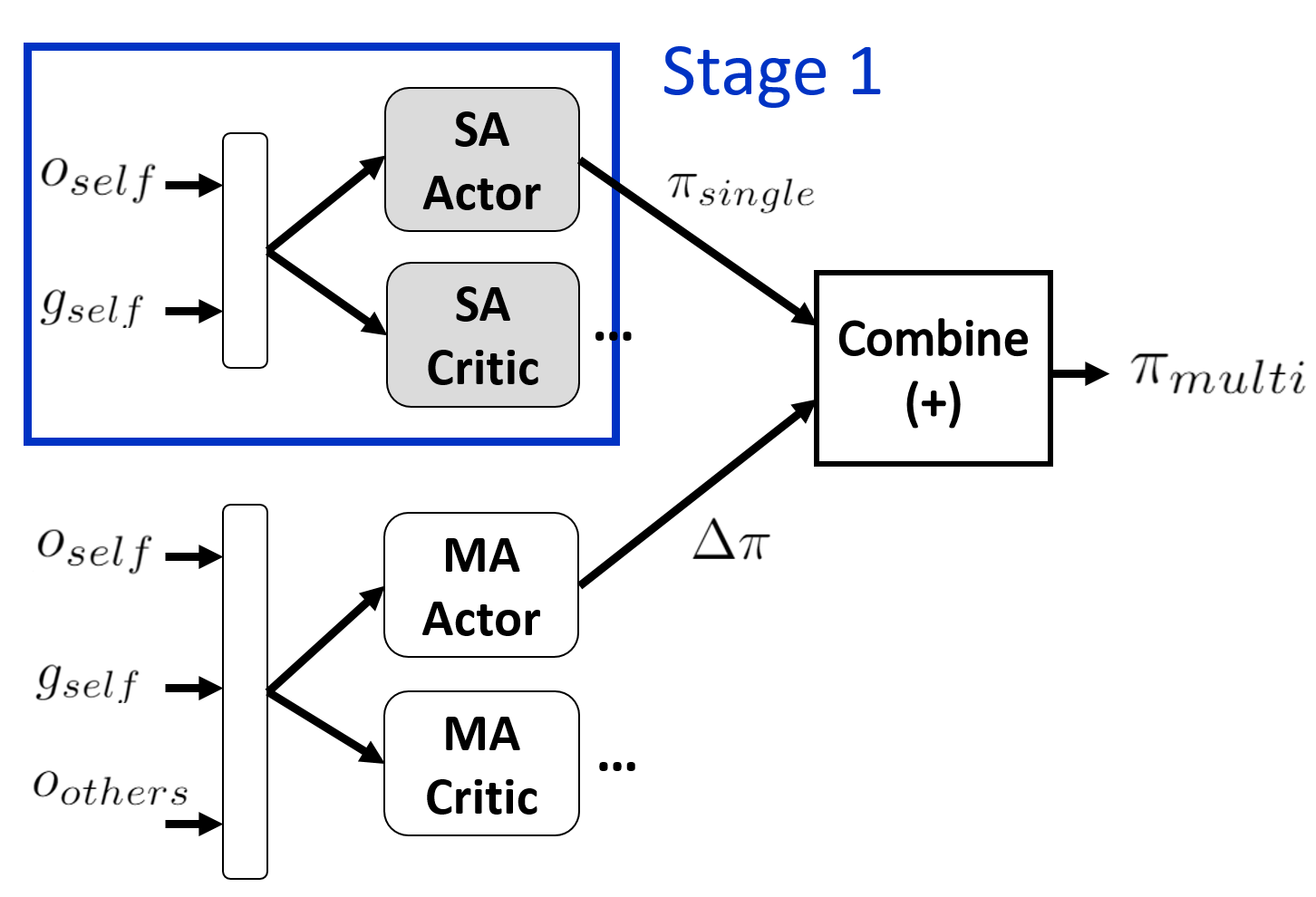}
  \caption{Multi-agent (MA) actor critic model. The gray colored boxes are the single agent policies and are frozen during training in the multi-agent setting.}
  \label{fig:multi_model}
  \vspace{-0.4cm}
\end{figure}

\noindent We assume that each agent has a noisy estimate of the other agents' states, but they don't have access to the other agents actions or intentions. 
We model the multi-agent decision problem as a Markov Game ~\cite{littman1994markov}, modified to accommodate mixed goals, and defined as the tuple $<N,\mathcal{S},{\{O_i\}}_{i \in N},{\{A_i\}}_{i \in N},{\{R_i\}}_{i \in N},{\{G_i\}}_{i \in N}, P, \mathcal{G},\gamma>$ with N agents. The possible configurations of the agents is specified by $\mathcal{S}$. Each agent $i$ gets an observation $o_i \in O_i$ which includes both the agent's observation of its own state $o_{s}$ and an observation of other agents $o_{o}$. Each agent has its own set of actions $A_i$, 
a goal $G_i \sim \mathcal{G}$,  and a reward function $R_i: \mathcal{S} \times \mathcal{G} \times A_i \rightarrow \mathbb{R}$. The markov game includes a transition function $P : \mathcal{S} \times A_1 \times ... \times A_N \rightarrow \mathcal{S}$ which determines the distribution over next states. The solution to the markov game is a policy for each agent $i$ $\pi_{\theta_i}: O_i \times \mathcal{G} \times A_i \rightarrow [0,1]$ where $\theta_i$ is the parameters of the policy. For continuous actions problems, $\pi_{\theta_i}$ is assumed to be a Gaussian where the mean is modeled by neural networks. Each agent seeks to find a policy $\pi_{\theta_i}$ that maximizes its own expected future discounted reward $R_i=\sum_{t=0}^{T} \gamma^{t}{{r_i}}_t$. In the following paragraphs, we remove $i$ for simplicity and add subscript ``self" or ``s" to our notation to show the agent's own properties. We add ``multi" or ``mlt'' to highlight the multi-agent scenario, and subscript ``others'' or ``o'' refers to other agents properties.

We modify each agent's $R_{sng}$ to account for the presence of other agents in the environment. Each agent is rewarded based on its individual objective, but is punished if it gets into conflicts (\emph{e.g.}, collisions in mobile agent scenarios) with other agents. The new reward function for each agent is as follows 
where $C$ is a positive constant that penalizes the agent for conflicts, and $\mathbbm{1}_{conflict}(s_{s},s_{o})$ determines if conflict is present:
\begin{align*}
\hspace{-8pt}
R_{mlt}(s_{s},a_{s},g_{s},s_{o}) = R_{sng}(s_{s},a_{s},g_{s}) - C \times \mathbbm{1}_{conflict}(s_{s},s_{o}) 
\end{align*}

We use a decentralized actor-critic policy gradient algorithm to learn the multi-agent policies. Each agent learns an actor-critic model that accounts for the multiple agents in the environment. The  model gets observation $o_{s}$, goal $g_{s}$, and $o_{o}$ as inputs, and outputs an action $a_{s}$ that the agent should take. Each agent learns a policy $\pi_{mlt}(a_{s}|o_{s},g_{s},o_{o})$, according to its multi-agent goal-specific reward function $R_{mlt}(s_{s},a_{s},g_{s},o_{o})$ in the presence of other agents. The decentralized actor-critic policy gradient algorithm maximizes $J_{mlt}(\theta)=\mathop{\mathbb{E}}_{o_{s} \sim p_{s}^\pi,a_{s} \sim {{\pi}_{mlt}},g_{s} \sim \mathcal{G},o_{o} \sim p_{o}^\pi}{[R_{mlt}]}$ by ascending the gradient:
\begin{multline*}
 \nabla_{\theta} {J(\theta)} = 
 \mathop{\mathbb{E}}_{\tau \sim p_{\theta}(\tau), g_{s} \sim \mathcal{G}} [\nabla_{\theta} \log {\pi_{mlt}(a_{s}|o_{s},g_{s},o_{o})} \\
A^{\pi}(o_{s},a_{s},g_{s},o_{s})]
\end{multline*}
Where $p_{o}^\pi$ is the other agents' state distribution. 
We use $\tau \sim p_{\theta}(\tau)$ to refer to $o_{s} \sim p_{s}^\pi,a_{s} \sim {{\pi}_{mlt}},o_{o} \sim p_{o}^\pi$.
Fig. \ref{fig:multi_model} shows our architecture for both reaching the individual goal and cooperative non-conflicting behavior. In this architecture, each agent leverages its learned single agent actor-critic models. The learned model is frozen and combined with another multi-agent model that addresses the cooperative non-conflicting behavior. 

The multi-agent module includes the single agent (SA) module from the previous stage of the curriculum. The output of the single agent model is combined with the output of a multi-agent (MA) module to learn a policy that modifies the single agent value functions to account for the other agents. The agent's own state $o_{s}$ and an estimation of the other agents state $o_{o}$ are passed to the actor and critic models of the multi-agent module. In Fig. \ref{fig:multi_model}, we only show the actor models for simplicity, the critic models have the same structure. In this work, we used a summation to combine the single agent and multi-agent models since we found it sufficient in our experiments. 


\begin{figure*}[tb]
    \centering
    \begin{subfigure}{0.45\textwidth}
    	\centering
        \includegraphics[width=0.6\textwidth]{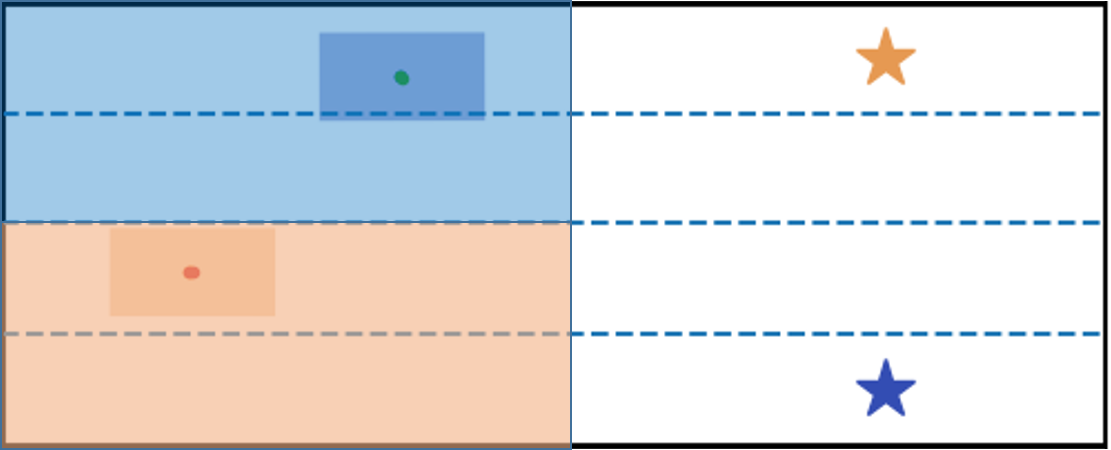}
        \caption{Lane-change environment. The bottom and top agents start from the bottom left and top left quarters respectively. }
        \label{fig:lane_reset}
    \end{subfigure}
    \begin{subfigure}{0.45\textwidth}
    	\centering
        \includegraphics[width=0.6\textwidth]{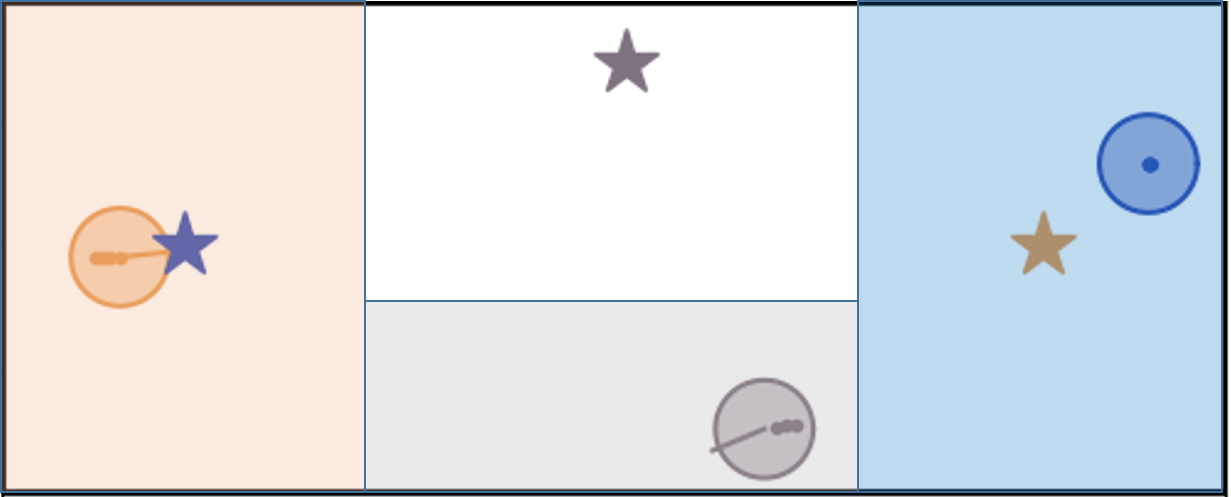}
        \caption{Robot navigation environment. Each agent starts from a random position in their corresponding side of the course.}
        \label{fig:robot_reset}
    \end{subfigure}
    \caption{Environments. Agents are crossing one another to go to their opposite side of the course.}\label{fig:envs}
    \vspace{-0.4cm}
\end{figure*}

\section{EXPERIMENTS}

\noindent In this section, first we discuss our network architecture. We then delve into the details of the environments and experimental setup, and then discuss our results.

\subsection{Algorithm and Network Architecture}
\noindent We use the Trust Region Policy Optimization (TRPO) \cite{schulman2015trust} algorithm to learn the actor-critic models. We use the same architecture and parameters as the OpenAI baseline implementation~\cite{baselines}. We use ReLU as the activation function instead of tanh in the original implementation. The actor-critic models each have 2 hidden layers with 128 neurons. We call our approach Interaction-Aware TRPO or IATRPO. 

\subsection{Simulation Environments}
\noindent We tested our proposed approach on the following two environments. Both environments are designed such that interaction is required for successfully achieving the individual goals. The environments are shown in Fig. \ref{fig:envs}.
\subsubsection{Lane-change}
\noindent This environment consists of two cars that are crossing to go to their goal destination (matching color stars). At each episode a pair of non-adjacent goals are selected and are kept constant throughout the episode. The agents' positions are selected randomly in the top left and bottom left quarters of the course. The agents start with a $0$ velocity, $0$ angular velocity, and $0$ heading angle. We call this environment ``C2". We created an easier version of this environment where the goals are fixed to the 1st and 3rd lanes, 
and the agents' position 
has randomness only in the $+x$ direction. We call this environment ``C2-fixed".

The agent's state includes its $x$ and $y$ position, velocity, angular velocity, heading angle, if it is broken due to collision with other agents or the environment, and if it has reached its goal. The observation noise for $s_{s}$ is in $[-0.01,0.01]$. The cars have acceleration and angular acceleration as their actions $a_{s}$ with uniform noise in $[-0.1,0.1]$. The car can reach a minimum and maximum velocity of $-1$ and $1$ respectively, and a minimum and maximum angular velocity of $-1$ and $1$ respectively. The multi-agent module for each agent has access to other agent's x and y positions, velocity, heading, angular velocity with a uniform noise in $[-0.1,0.1]$. The cars use a bicycle kinematics model~\cite{kong2015kinematic}.

The reward function for the single agent scenario and the multi-agents scenario are as follows with a reward scale $3$.  Function $d$ computes the euclidean distance between the center of the car and agent's goal. Function $collision(s_s,env)$ or $collision(s_s,s_o)$ specify if the agent is in collision with the environment or the other agents respectively.
\[
    R_{sng}(s_{s},a_{s},g_{s}) =
\begin{cases}
    -1, & \text{if }collision(s_s,env)\\
     1,              & \text{if }d(s_s,g_s) < 0.4\\
    \frac{d(s_s,g_s)}{1000},              & \text{otherwise}
\end{cases}
\]
\[
    R_{mlt}(s_{s},a_{s},g_{s},s_{o}) =
\begin{cases}
	-1, & \text{if }collision(s_s,env)\\
    -1, & \text{if }collision(s_s,s_o)\\
     1,              & \text{if }d(s_s,g_s) < 0.4\\
    \frac{d(s_s,g_s)}{1000},              & \text{otherwise}
\end{cases}
\]

\begin{figure*}[tb]
  \centering
	\noindent
	\filmbox{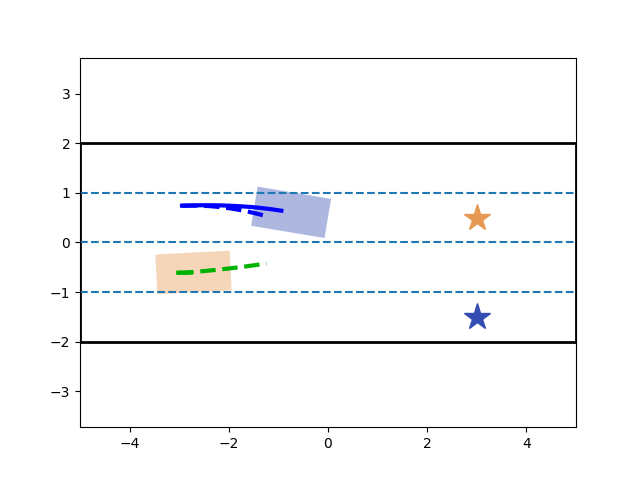}\filmbox{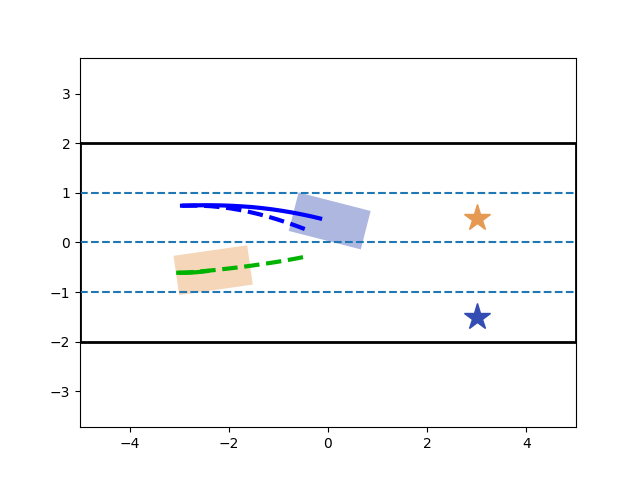}\filmbox{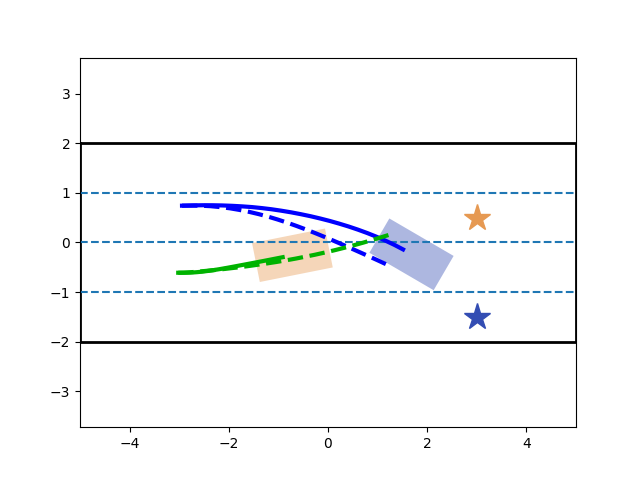}\filmbox{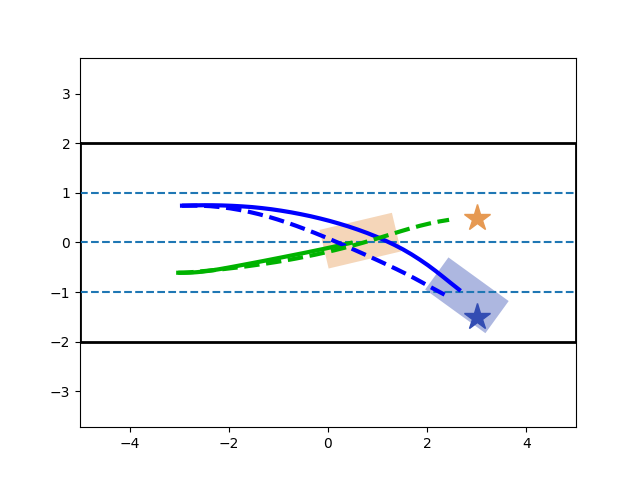}\filmbox{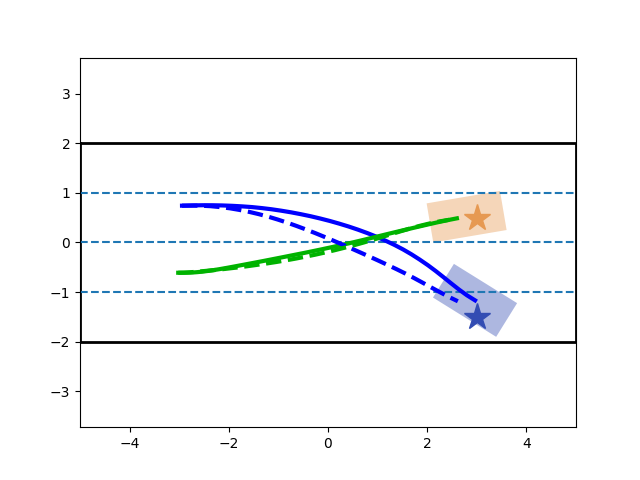}
	\caption{IATRPO's final policy on the lane-change environment.}
  \label{fig:filmstrip_car}
\end{figure*}

\begin{figure*}[tb]
  \centering
	\noindent
	\filmbox{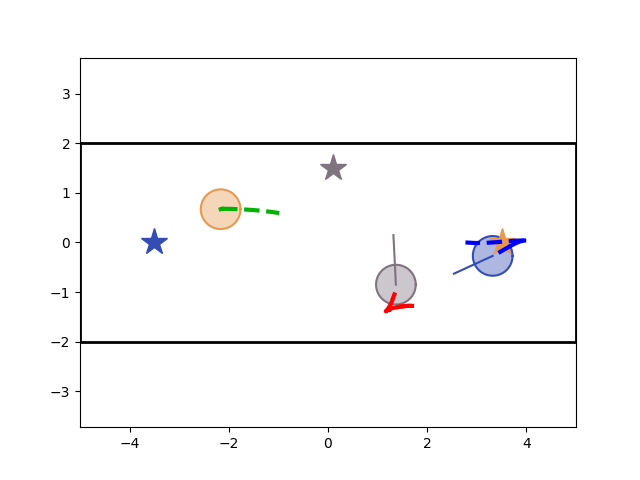}\filmbox{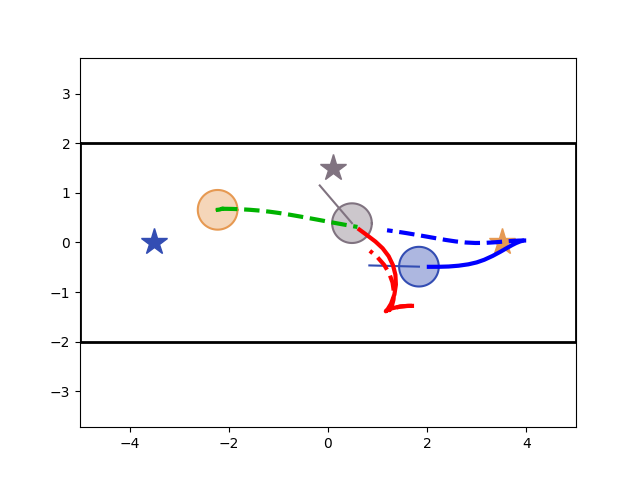}\filmbox{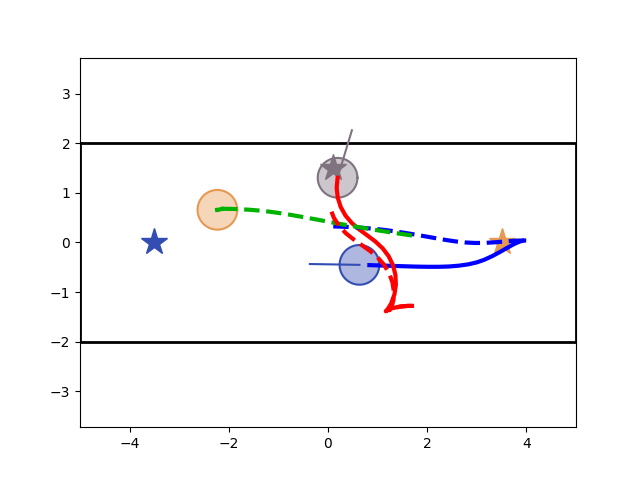}\filmbox{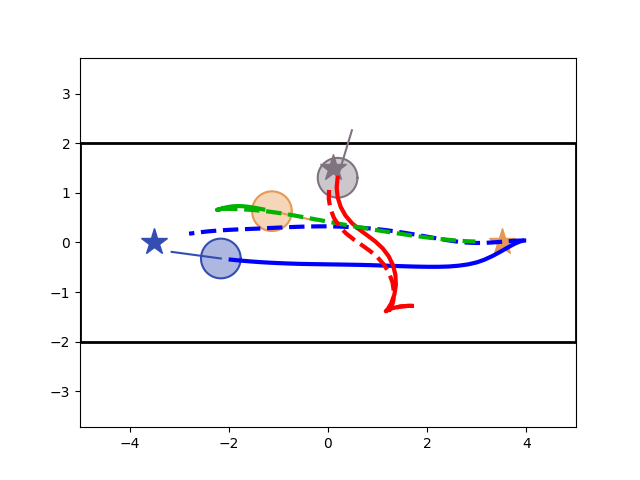}\filmbox{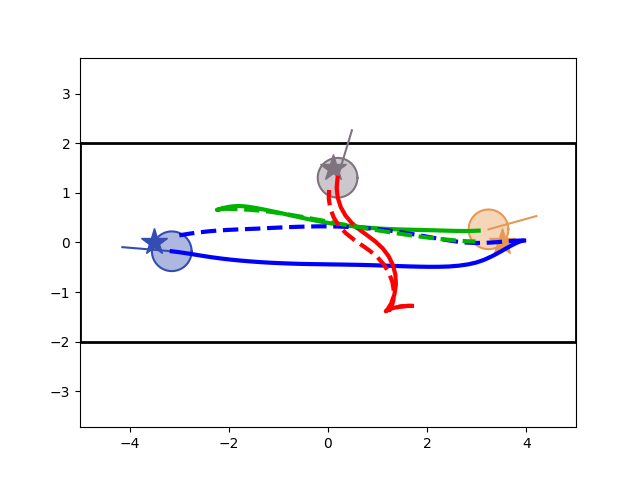}
	\caption{IATRPO's final policy on the robot navigation environment.}
  \label{fig:filmstrip_robot}
  \vspace{-0.3cm}
\end{figure*}


\subsubsection{Multi-Robot Navigation}

\noindent This environment consists of two or more mobile robots that are crossing one another to go to their goal destination (matching color stars). Three fixed goals are located at the top, left and right sides of the course. The agents' positions are selected randomly in the left (goal in right region of the environment), right (goal in left region), and bottom (goal in top region) of the course to assure that the agents pass one another to go to their goal position. The 2 agent environment has the same setting, but the bottom (gray) agent is not present. We call the environment with 2 agents and 3 agents ``R2'' and ``R3'' respectively.
The robots have the same state space and parameters as the cars in the lane changing environment. The mobile robots use the unicycle kinematics model~\cite{indiveri1999kinematic}. The reward function is as before.  


\subsection{Results}
\noindent We provide quantitative and qualitative results on the performance of our approach. As our baseline, we compare against MADDPG~\cite{lowe2017multi}.  
We leveraged the main contribution of the MADDPG approach and  implemented the multi-agent version of the TRPO~\cite{schulman2015trust} algorithm (MATRPO) where each agent is provided with the action values of the other agents. TRPO was used in place of DDPG because it was found to consistently outperform DDPG in all our experiments. TRPO was also used to train the IATRPO for all stages of the curriculum. 


\subsection{Qualitative Results}

\noindent Figures \ref{fig:filmstrip_car} and \ref{fig:filmstrip_robot} illustrate the agents learned policies on the two environments. In both scenarios, all the agents must cross paths to reach their goal destination (matching color stars). We show the single agent policies with dashed lines, and the multi-agent policies with solid lines. We ran the learned multi-agent models and observed that the agents are successfully able to learn how to interact. We refer to the agents based on their colors or their start positions.

In the multi-agent policy in Fig. \ref{fig:filmstrip_car}, the bottom agent (orange or abbreviated as O) slows down for the top agent (blue or B) to pass first and then it goes to its goals (but the single agent trajectory is in collision with the other agent). The top agent (B) also modifies the shape of its trajectory to not get into collision with the bottom agent. In the multi-agent policy in Fig.  \ref{fig:filmstrip_robot}, the bottom agent (gray or G) learns to go first with maximum speed, the right agent (B) slows down and modifies the shape of its trajectory for the bottom agent (G) to pass first. The left agent (O) modifies its speed to prevent a collision with the other agents. Notice that the single agent policies differ both in speed and shape from the multi-agent policies, and if all the agents executed their single agent policy, they would have collided with the others. Please refer to the video accompanying the paper to see examples of successful and failed executions.

\begin{table}[tb]
\caption{Success of the algorithms on the 4 environments.}
\centering
\begin{center}
\begin{tabular}{|c || c || c |}
\hline
  Environment & MATRPO success rate (\%) & IATRPO success rate (\%) \\
    \hline
    {C2-fixed} & $97.88\pm0.31$ & $99.4 \pm 0.33$\\
    \hline
    {C2}& $0 \pm 0$ & $94.3 \pm 2.99$\\
    \hline
    {R2}& $36.92 \pm 45.22$ & $90.96 \pm 1.43$\\
    \hline
    {R3}& $0 \pm 0$ & $88.02 \pm 4.11$\\
    \hline
\end{tabular}
\end{center}
\label{tab:success}
\vspace{-0.6cm}
\end{table}

\subsection{Quantitative Results}


\noindent We evaluate the IATRPO approach against the MATRPO approach and report the final results in three evaluations:

\parskip = 8pt
\noindent\textbf{Success Rate:}
Table \ref{tab:success} shows the success rate of our approach against the MATRPO approach. We ran both approaches on the 4 environments with 5 random seeds. An episode is successful if both agents are $0.4$ away from their goals ${d}(s_{self},g_{self}) < 0.4$. To compute the success rate, the final learned policies were run on $1000$ random episodes. Both MATRPO and IATRPO approaches give a high accuracy on the C2-fixed environment, but the success rate of the IATRPO algorithm is higher. On the C2 environment, which has a greater amount of randomness in the start position and has random goals, the MATRPO algorithm is not able to learn successful policies for all the agents, but the IATRPO algorithm has $94.3\pm2.99 \%$ success rate. 
MATRPO learns a successful policy for one agent, but the second agent is stuck in a local optima, having only learned to not collide with the environment or the successful agent. Most of the failure cases in IATRPO happens around the boundaries of the environments or when the agents are too close to one another. We believe this is because of the noise associated with both the agent's action and the observations of others.
\parskip = 0pt

\begin{figure} [tb]
    \centering
    \begin{subfigure}[b]{0.2\textwidth}
        \includegraphics[width=\textwidth,trim={0.6cm 0.6cm 0.6cm 0.1cm},clip]{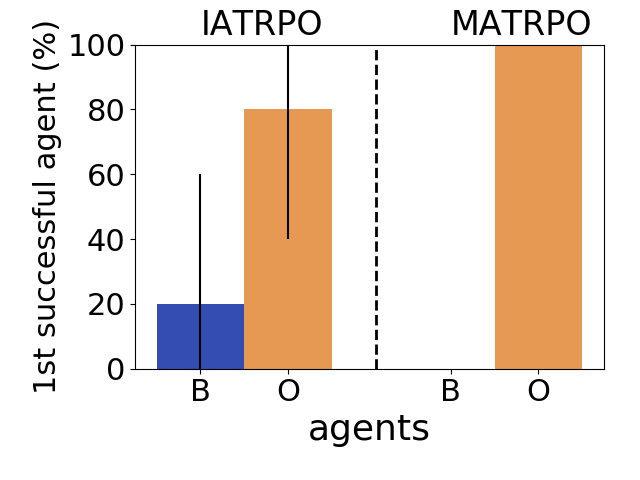}
        \caption{Results on C2-fixed.}
        \label{fig:iatrpo_train}
    \end{subfigure}
    \begin{subfigure}[b]{0.2\textwidth}
        \includegraphics[width=\textwidth,trim={0.6cm 0.6cm 0.6cm 0.1cm},clip]{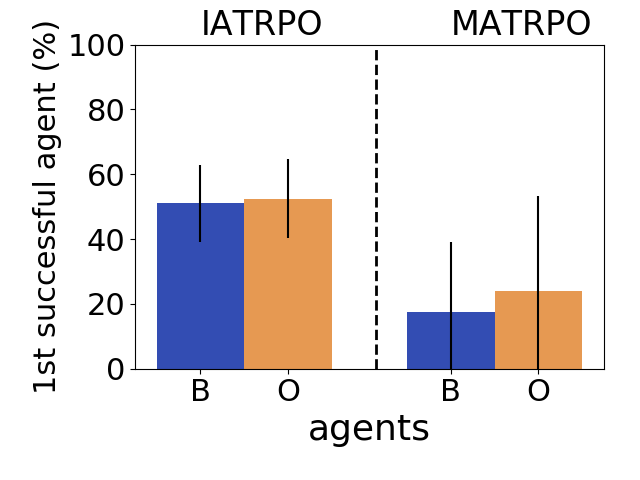}
        \caption{Results on R2.}
        \label{fig:matrpo_train}
    \end{subfigure}
    \caption{Shows which agent achieved its goal first. }
    \label{fig:first}
    \vspace{-0.4cm}
\end{figure}

\begin{figure} [tb]
    \centering
    \begin{subfigure}{0.235\textwidth}
        \includegraphics[width=\textwidth,trim={0.6cm 0.6cm 0.6cm 0.6cm},clip]{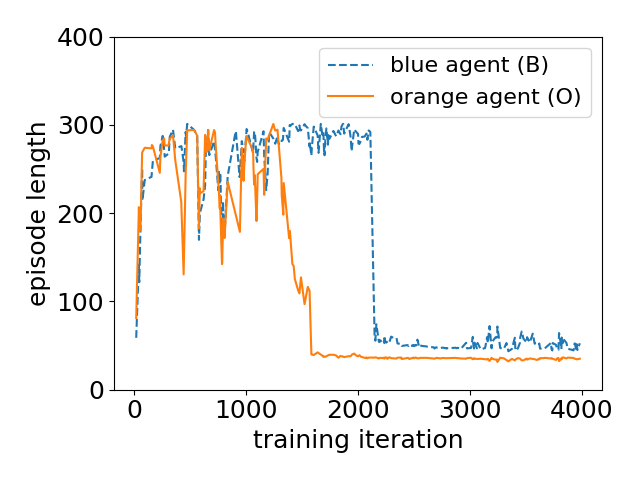}
        \caption{MATRPO training.}
        \label{fig:matrpo_train}
    \end{subfigure}
    \begin{subfigure}{0.235\textwidth}
        \includegraphics[width=\textwidth,trim={0.6cm 0.6cm 0.6cm 0.6cm},clip]{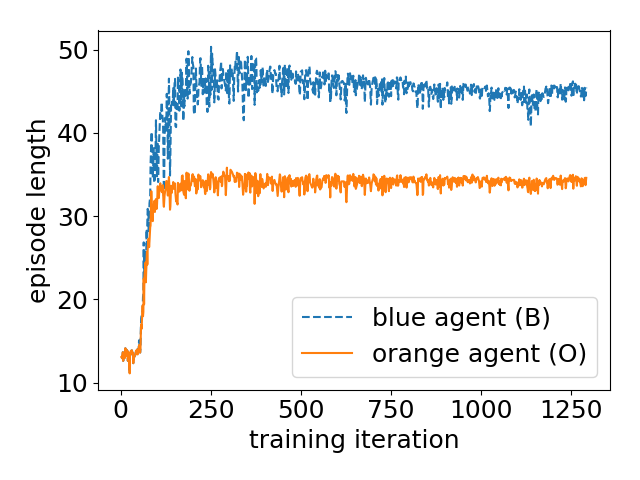}
        \caption{IATRPO training.}
        \label{fig:iatrpo_train}
    \end{subfigure}    
    \caption{Mean episode length in one experiment for training multi-agent policies on the C2-fixed. IATRPO uses a curriculum so the number of iterations is not comparable.}
    \label{fig:train}
    \vspace{-0.4cm}
\end{figure}

The performance of the IATRPO algorithm is much higher than the MATRPO algorithm on the R2 environment. In half of the random runs the MATRPO algorithm did not learn a successful policy for both the agents. However, the IATRPO algorithm has a success rate of around $90\%$. The MATRPO algorithm completely failed to learn a successful policy on the R3 algorithm, but the IATRPO algorithm achieves a success rate of $88.02\pm4.11 \%$.  
\\
\\
\noindent\textbf{Level of Interaction:}
We measure how interactive are the final policies that are learned by the IATRPO and MATRPO approaches. We ran both approaches on C2-fixed and R2 environments where MATRPO was able to learn successful policies. We performed 5 training runs with random seeds and tested the final learned policies on a $1000$ random episodes. We estimate how interactive the policies are by finding which agent reached its goal first. For each agent, we compute the average and standard deviation of the interactiveness metric on the 5 runs. In each algorithm run if one agent always waits for the other agent to go first and compromises, the agents are considered non-interactive. In a two agents scenario, the ideal case is if both agents have an average of around $50\%$ with a low standard deviation. Fig. \ref{fig:first} shows the results of the two algorithms on the two environments. In each run on the C2-fixed environment where we applied the MATRPO algorithm, one agent always waited for the other agent to go first. However, the IATRPO algorithm was able to learn more interactive policies than the MATRPO policies where both the agents sometimes compromised. Fig. \ref{fig:train} provides more evidence of why, in the MATRPO training, one agent always compromises.

Fig. \ref{fig:train} shows the mean episode length of both the algorithms in one of the training runs on the C2-fixed environment. When the mean episode length becomes constant, the agent has converged to a successful policy. In the MATRPO training, the orange agent converges to the successful policy and after about 800 training iterations the blue agent adapts its policy to the orange agent's policy and converges as well. However, with IATRPO the two agents learn a successful policy around the same time.

We applied both the algorithms on the R2 environment and noticed that in the IATRPO algorithm the two agents have a better balance where both the agents achieve the first place about $50\%$ of the times. The agents are less balanced when using the MATRPO approach and have higher variance than the IATRPO approach. 
\begin{figure}[tb]
    \centering
    \begin{subfigure}{0.2\textwidth}
        \includegraphics[width=\textwidth,trim={0.8cm 0.8cm 0.8cm 0.5cm},clip]{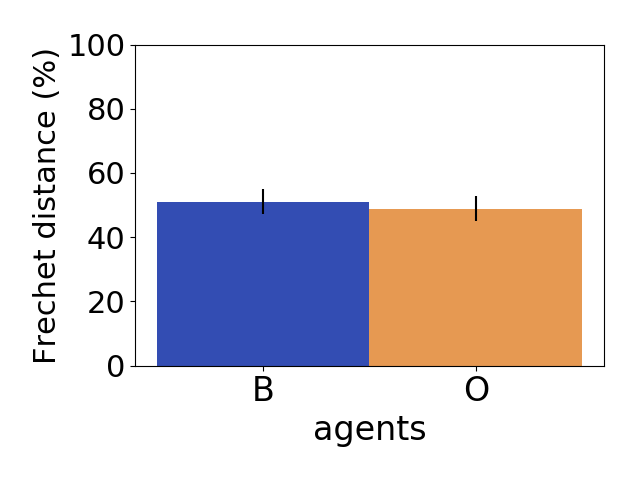}
        \caption{C2 environment.}
        \label{fig:C2_trajs_dist}
    \end{subfigure}
    \begin{subfigure}{0.2\textwidth}
        \includegraphics[width=\textwidth,trim={0.8cm 0.8cm 0.8cm 0.5cm},clip]{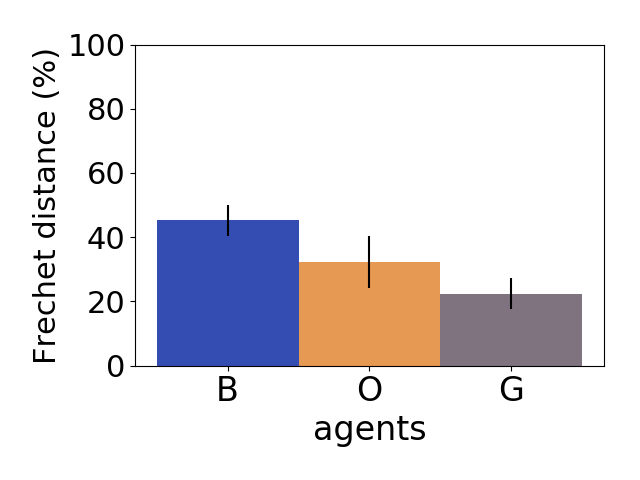}
        \caption{R3 environment.}
        \label{fig:R3_trajs_dist}
    \end{subfigure}
    \caption{Fr\'echet distance between the single agent trajectories and the multi-agent trajectories in IATRPO algorithm.}
    \label{fig:trajs_dist}
     \vspace{-0.6cm}
\end{figure}
We also investigate the influence of our two stages approach on the interactiveness of the agents. We measured the distance between the single agent trajectories and the multi-agent trajectories to measure how much each agent modified its trajectory to account for the other agents in the IATRPO algorithm. We use the Fr\'echet distance for our comparison. As before, we use the final learned policy, run it on $1000$ random episodes and compute the distance between the single agent module's trajectory and the multi-agent module's trajectory. For each agent, we average the computed distance and use that to compute the overall compromise ($\%$) that each agent makes compared to other agents. Fig. \ref{fig:trajs_dist} shows the results on the C2 and R3 environments. Although the top agent gets the first place $72.43 \pm 37.19\%$ of the times, the changes in the distance is almost equal for the two agents in the C2 environment. 

In the R3 environment, the bottom agent (G) always arrives first at the goal, but the distance between its single agent trajectory and multi-agent trajectory is about $20\%$. This implies that the bottom agent is also trying to adapt its policy to the other agents' policies. The right agent and the left agent get the second place $82.59 \pm 22.84\%$ and $18.46 \pm 22.64\%$ respectively. The overall impact of the right agent (B) on distance is $45\%$ compared to the left agent (O) $32\%$. This implies that both agents change their single agent policies, the right agent mostly changes the shape of its policy and the left agent mostly changes its speed to account for the other agents.
\\
\\
\noindent\textbf{Mixed Agents:}

 \begin{table}[tb]
\caption{Success rate of the algorithms on the 4 environments when tested on agents not trained together.}
\centering
\begin{center}
\begin{tabular}{|c || c || c |}
\hline
  Environment & MATRPO success rate (\%) & IATRPO success rate (\%) \\
    \hline
    {C2-fixed} & $97.83\pm0.72$ & $98.71\pm1.29$\\
    \hline
    {C2}& NA & $69.22 \pm 26.44$\\
    \hline
    {R2}& $2.18 \pm 8.56$ & $77.05 \pm 9.46$\\
    \hline
    {R3}& NA & $68.87 \pm 17.44$\\
    \hline
\end{tabular}
\end{center}
\label{tab:success_pairs}
\vspace{-0.4cm}
\end{table}

\noindent We conducted experiments where we look at the performance of agents not trained together. This is a scenario known in the literature to cause agents to fail due to dependent policies ~\cite{raghu2017can,lanctot2017unified}. We used the 5 random training runs and generated 20 pairs (or triples) of agents where the agents in each pair are trained separately using a different seed. Table~\ref{tab:success_pairs} shows the success rate of the MATRPO and IATRPO algorithms on the four environments. We used the same approach as above to compute the success rate on a 1000 random episodes. The performance of the MATRPO algorithm is not affected much in C2-fixed experiments, but it has drastically decreased in R2 experiments. We believe the reason for this behavior is the following. In C2-fixed experiments with MATRPO, the bottom agent (O) learns to always go first regardless of what the top agent (B) is doing. Even when the bottom agent is tested against other top agents (Bs), both the agents show the same behavior thus the performance of the algorithm does not get affected. However, in the R2 experiments with MATRPO, the agents show a more interactive behavior than C2-fixed thus when we test the agents, which were not trained together, against each other, the performance drastically decreases.

The success rate of the IATRPO algorithm also decreases when we test it on the 20 pairs (or triples). The performance degrades less on the easier environments such as C2-fixed and on the environments where the agents learned a more interactive policy with low variance such as R2. The success rate for the C2 and R3 degrades more than R2 since the agents learned a less interactive behavior with high variance.

\section{CONCLUSIONS}

\noindent We focus on multi-agent settings where each agent learns a policy to simultaneously achieve its individual goal and interact with others. We provide a curriculum learning approach and a  architecture that learn how to adapt single agent policies to the multi-agent setting. We tested the method on two robotics problems and observed that our approach outperforms the state-of-the-art approach and results in interactive policies. 

Our formulation is generalizable to domains with inhomogeneous agents since we make no assumptions regarding the homogeneity of the agents, and our future work involves testing the approach on such domains. 
One method that we are planning to try is the ensemble of policies method proposed in~\cite{bansal2017emergent}.
If a model is learned in an environment with $N$ agents, we can apply the same model on an environment with $<=N$ agents where we assume the non-existent agent is in a corner and is not interacting with others. However, a limitation of our work is that a new model should be learned if we increase the number of agents. We believe this issue can be addressed by leveraging an approach that is agnostic to the number of agents such as~\cite{schwab2018zero}.

\bibliographystyle{plain}
\bibliography{IEEEabrv,references} 
\end{document}